\documentclass{ecai} 

\usepackage{latexsym}
\usepackage{amssymb}
\usepackage{amsmath}
\usepackage{amsthm}
\usepackage{booktabs}
\usepackage{enumitem}
\usepackage{graphicx}
\usepackage{color}
\usepackage{multirow} 
\usepackage{threeparttable}

\newcommand{\BibTeX}{B\kern-.05em{\sc i\kern-.025em b}\kern-.08em\TeX}

\begin{document}


\begin{frontmatter}


\paperid{123} 


\title{MCLPD: Multi-view Contrastive Learning for EEG-based PD Detection Across Datasets}


\author[A,B]{\fnms{Qian}~\snm{Zhang}}
\author[A,B]{\fnms{Ruilin}~\snm{Zhang}}
\author[A,B]{\fnms{Jun}~\snm{Xiao}}
\author[A,B]{\fnms{Yifan}~\snm{Liu}}
\author[A,B]{\fnms{Zhe}~\snm{Wang}\thanks{Corresponding Author. Email: wangzhe@ecust.edu.cn}}
%

\address[A]{Key Laboratory of Smart Manufacturing in Energy Chemical Process, Ministry of Education, Shanghai, P. R. China}
\address[B]{East China University of Science and Technology, Shanghai, P. R. China}


\begin{abstract}

Electroencephalography has been validated as an effective technique for detecting Parkinson’s disease, particularly in its early stages.
However, the high cost of EEG data annotation often results in limited dataset size and considerable discrepancies across datasets, including differences in acquisition protocols and subject demographics, significantly hinder the robustness and generalizability of models in cross-dataset detection scenarios.
To address such challenges, this paper proposes a semi-supervised learning framework named MCLPD, which integrates multi-view contrastive pre-training with lightweight supervised fine-tuning to enhance cross-dataset PD detection performance.
During pre-training, MCLPD uses self-supervised learning on the unlabeled UNM dataset. 
To build contrastive pairs, it applies dual augmentations in both time and frequency domains, which enrich the data and naturally fuse time-frequency information.
In the fine-tuning phase, only a small proportion of labeled data from another two datasets (UI and UC) is used for supervised optimization. Experimental results show that MCLPD achieves F1 scores of 0.91 on UI and 0.81 on UC using only 1\% of labeled data, which further improve to 0.97 and 0.87, respectively, when 5\% of labeled data is used.
Compared to existing methods, MCLPD substantially improves cross-dataset generalization while reducing the dependency on labeled data, demonstrating the effectiveness of the proposed framework.

\end{abstract}

\end{frontmatter}


\section{Introduction}

Parkinson’s disease (PD) is a chronic neurodegenerative disorder characterized primarily by motor impairments, including resting tremor, muscular rigidity, bradykinesia, and postural instability, which significantly deteriorate patients’ quality of life \cite{c:3,c:4}.
Although pharmacological treatments and deep brain stimulation (DBS) have achieved progress in symptom management recent years, they remain ineffective in halting disease progression. Consequently, early diagnosis has become a critical focus in PD research.
Electroencephalography (EEG), as a non-invasive brain monitoring technique \cite{c:1,c:2}, enables the recording of brain electrical activity and reveals frequency-specific abnormalities in patients with neurological disorders.
Thus, EEG plays a vital role in the auxiliary early diagnosis and treatment evaluation of PD.

In recent years, deep learning has been explored and applied to EEG analysis.
Early studies on PD detection are primarily conducted in single-dataset settings \cite{c:7, c:17, c:18}.
However, due to the variations in data acquisition protocols and subject characteristics across different medical institutions, as well as the limited size of each dataset, models trained on one single dataset often fail to generalize effectively to real-world clinical environments.
This limitation highlights the necessity of developing methods capable of handling cross-dataset scenarios.
Existing approaches for cross-dataset PD detection face challenges.
Some works rely on handcrafted features, which are susceptible to distributional shifts and exhibit limited generalization ability \cite{c:6,c:28,c:29}.
Others incorporate prototype learning to enhance robustness \cite{c:5}, but remain highly dependent on sample quality and feature design.
Therefore, it is imperative to develop methods that can learn deep representations from EEG data with minimal reliance on labeled samples.
To this end, self-supervised learning has emerged as a promising direction.
Among these methods, contrastive learning operates by comparing positive and negative sample pairs, aiming to increase the similarity between different views of the same instance while reducing the similarity between different instances.
This approach has been proven effective in various EEG-related tasks, including sleep staging \cite{c:11}, seizure prediction \cite{c:12}, and tinnitus diagnosis \cite{c:13}. However, these methods have not been validated in multi-dataset scenarios and have yet to be explored in the context of cross-dataset PD detection.

To address the above challenges, we propose MCLPD, a semi-supervised learning framework that integrates multi-view contrastive pre-training with supervised fine-tuning. 
During the pre-training phase, we conducts self-supervised representation learning on the unlabeled dataset UNM. To support the construction of contrastive sample pairs, we employ dual augmentation strategies in both time and frequency domains.
These strategies not only diversify samples but also inherently achieve time-frequency information fusion.
A dynamic augmentation manager model is designed to adaptively select the most effective augmentations.
By optimizing a contrastive loss, the model learns to align representations of the same signal under different augmented views, thereby obtaining more discriminative feature representations.
We evaluate the pretrained model on another two datasets (UI and UC datasets) without any fine-tuning.
The results show F1 scores of 0.75 on both datasets, demonstrating that the model has learned generalizable EEG representations.
Subsequently, we perform fine-tuning using only 1\% and 5\% of labeled data.
To further reduce label dependence and enhance generalization, we apply dynamic augmentation to fine-tuning data, along with partial encoder layer freezing, Lookahead optimizer, and Stochastic Weight Averaging (SWA).
Experimental results show that with just 1\% labeled data, the model achieves F1 scores of 0.91 and 0.81 on the UI and UC datasets, respectively.
With 5\% labeled data, the F1 scores further increase to 0.97 and 0.87.
These results outperform existing baselines in the scenario of across datasets, while significantly reducing training cost.
In addition, we conduct a perturbation-based multi-dimensional interpretability analysis.
Our findings offer potential new biomarkers for the development of non-invasive early diagnosis and monitoring tools in clinical practice, facilitating the optimization of therapeutic intervention timing.

The main contributions of this work are summarized as follows:

\begin{itemize} 
	\item 
	We propose a semi-supervised multi-view contrastive learning framework for EEG-based Parkinson's disease detection, integrating both pre-training and fine-tuning stages to enhance detection performance in cross-dataset scenarios.
	
	\item We incorporate both temporal and spectral information of EEG signals and introduce a dynamic data augmentation strategy that enhances feature diversity and strengthens model robustness.
	
	\item We conduct evaluation experiment and the proposed MCLPD achieves promising cross-dataset transferability. It reaches baseline-level performance with only 1\% of labeled data, and further achieves an F1 score of 0.97 with 5\% supervision, significantly outperforming existing baselines.
\end{itemize}


\section{Related Work}
This section introduces two key areas related to our study: EEG-based PD classification and contrastive learning in EEG.

\textbf{EEG-based PD classification}: In terms of PD detection, early studies employ traditional machine learning methods using features such as power spectral density (PSD) and Hjorth parameters for classification tasks \cite{c:16,c:15}.
More recently, deep neural networks have been applied to reduce the reliance on handcrafted features and achieve improved classification performance.
One of the earliest approaches adopt a multi-layer convolutional neural network (CNN) architecture \cite{c:17}.
Subsequent research proposes deeper and more advanced models, including a 20-layer CNN designed to differentiate between medicated and non-medicated PD patients \cite{c:18}.
Other studies introduce specialized architectures such as ASGCNN, which incorporates attention-based modules into a graph neural network (GNN) \cite{c:19}.
MCPNet further advances this direction by integrating multi-scale convolutional prototypes to enhance feature diversity and improve generalization \cite{c:5}.

\textbf{Contrastive learning in EEG}: Contrastive learning has shown promising results in various EEG-related tasks beyond PD detection.
For emotion recognition, Gilakjani and Osman \cite{c:20} combine contrastive learning with graph neural networks to capture both spatial and temporal dynamics, improving cross-subject classification accuracy.
In the context of epileptic seizure prediction, CLEP  \cite{c:13}  integrates a spatio-temporal spectral encoder with contrastive learning to reduce inter-subject variability and enhance model generalization.
Wang et al. \cite{c:14} introduce MECRL, which leverages multi-band EEG signals and multi-task learning within a contrastive framework to improve adaptability in tinnitus diagnosis.
For sleep staging, Jiang et al. \cite{c:11} apply contrastive learning by constructing positive pairs through data transformation, enabling self-supervised learning under limited labeled data conditions.
Furthermore, Xu et al. \cite{c:21} explore contrastive learning in unsupervised domain adaptation for cross-dataset EEG classification. This method jointly optimizes the feature distributions of the source and target domains to improve classification performance in novel environments.

Although contrastive learning has been explored in several EEG applications, its application to PD detection remains limited.
Most existing methods for PD detection are either validated on a single dataset or rely on large amounts of high-quality labeled data, making it difficult to achieve stable generalization in cross-dataset scenarios with limited labeled samples.
In contrast, our semi-supervised learning framework demonstrates superior performance and robustness under cross-dataset settings.

\begin{figure*}[h]
	\centering
	\includegraphics[width=1.0\textwidth]{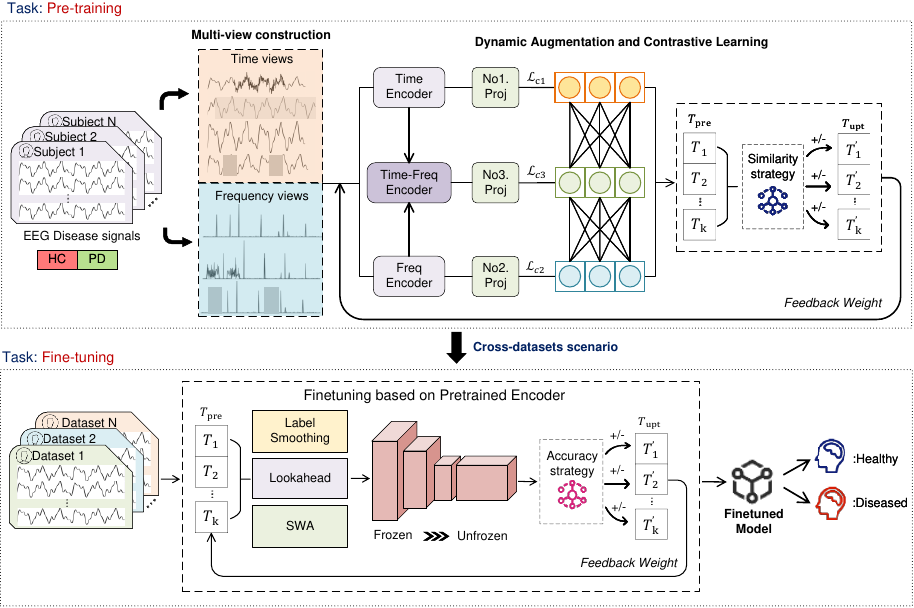}
	\caption{Overview of MCLPD's ability to detect Parkinson's disease with multi-view contrastive learning.}
	\label{fig1}
\end{figure*}

\section{Methods}

The proposed semi-supervised detection framework based on multi-view contrastive learning is illustrated in Figure \ref{fig1}.
In pre-training phase, the framework utilizes a time-frequency signal encoder with a dynamic data augmentation module, which adaptively selects effective augmentation strategies based on training feedback.
During fine-tuning on downstream datasets, the framework incorporates advanced optimization techniques including progressive layer unfreezing, label smoothing, Lookahead optimization, and Stochastic Weight Averaging (SWA).
A detailed description of each component is provided in the following.

\subsection{Dynamic Data Augmentation}
\subsubsection{EEG-signal Augmentation}
In contrastive learning, data augmentation plays a crucial role.
However, a single fixed augmentation strategy may fail to capture the intrinsic diversity of EEG signals, potentially leading to overfitting or limited generalization during training.
To address this issue, we employ a diverse set of augmentation techniques to generate multiple views of EEG signals across both time and frequency domains.

The time-domain augmentation methods that have been applied are detailed below.

%
%
%
%
%
%
%

\textbf{Gaussian noise:} Zero-mean Gaussian noise with variance \(\sigma^2\) is added to the original EEG signals \(S(t),\ t = 1, 2, \ldots, L\), where \(\sigma \in [0.05, 0.2]\). The noisy signal is defined as \(S'(t) = S(t) + N(t)\), where \(N(t) \sim \mathcal{N}(0, \sigma^2)\), introducing stochastic perturbations to enhance model robustness.

\textbf{Time shift:} EEG signals are circularly shifted along the temporal axis by a random offset \(\delta \in [-\delta_{\max}, \delta_{\max}]\), producing \(S'(t) = S((t - \delta) \bmod L)\), which simulates timing misalignments and enhances temporal invariance.

\textbf{Amplitude scaling:} The amplitude of EEG signals is scaled by a random factor \(\alpha \in [0.8, 1.2]\), resulting in \(S'(t) = \alpha \cdot S(t)\). This transformation maintains the waveform shape while modeling amplitude variation due to factors such as electrode impedance.

\textbf{Random masking:} A continuous segment \([t_s, t_e]\) is randomly selected and set to zero, i.e., \(S'(t) = 0\) for \(t \in [t_s, t_e]\) and \(S'(t) = S(t)\) otherwise, to simulate missing or corrupted data segments.

The frequency-domain augmentation methods that have been applied are detailed below.

\textbf{Frequency shift:} EEG signals are transformed using Fourier transform, and a random phase shift \(\Delta f\) is applied, yielding \(S'(t) = \mathcal{F}^{-1} \left( |\mathcal{F}(S(t))| \cdot e^{j(\phi + \Delta f)} \right)\), where \(\phi\) is the original phase and \(\mathcal{F}^{-1}\) denotes the inverse transform.

\textbf{Spectral scaling:} The spectral amplitude is modified by a scaling factor \(\beta \in [0.5, 1.5]\), resulting in \(S'(t) = \mathcal{F}^{-1} \left( \beta \cdot |\mathcal{F}(S(t))| \cdot e^{j\phi} \right)\), where \(\phi\) is the original phase. This adjusts frequency band energy while preserving structure.

\textbf{Band noise:} Frequency-domain noise is introduced by applying a binary mask \(M(f)\) to add complex Gaussian noise \(Z(f) \sim \mathcal{CN}(0, \sigma^2)\) to selected bands and simulate localized spectral perturbations, resulting \(S'(t) = \mathcal{F}^{-1} \left( \mathcal{F}(S(t)) + M(f) \cdot Z(f) \right)\).

\subsubsection{Dynamic Sampling Strategy}

During model training, the demand for data augmentation strategies evolves alongside the model's learning status.
Fixed augmentation combinations often fail to adapt to these dynamic requirements. To address this issue, we propose a dynamic augmentation manager, which adaptively selects the optimal augmentation strategy combinations based on feedback from the model.
Let $\mathcal{T} = \{T_1, T_2, \ldots, T_K\}$ denote the set of all available augmentation operators. We assign each operator $T_i$ a success score $s_i$ that reflects its effectiveness, and define the probability $p_i$ of selecting $T_i$ as follows:
\begin{eqnarray}
	p_i &=& \frac{\exp\left(\frac{s_i}{T}\right)}{\sum_{j=1}^{K} \exp\left(\frac{s_j}{T}\right)},
\end{eqnarray}
where $T$ is a temperature parameter that controls the smoothness of the probability distribution. The success score $s_i$ is computed as:
\begin{eqnarray}
	s_i &=& \frac{n_i^{\text{success}}}{n_i^{\text{total}}},
\end{eqnarray}
where $n_i^{\text{success}}$ denotes the number of successful applications of $T_i$, and $n_i^{\text{total}}$ is the total number of times it has been applied.

Based on the computed probabilities $\{p_i\}$, a subset of augmentation operators is randomly sampled and sequentially composed into a joint augmentation function \(T_{\text{combo}}(\mathbf{X}) = T_{i_n} \circ T_{i_{n-1}} \circ \cdots \circ T_{i_1}(\mathbf{X})\).
We define stage-specific criteria for determining whether an augmentation operation is successful.
In the pre-training phase, where the goal is to facilitate contrastive learning, we evaluate augmentation success based on the cosine similarity between the original representation $\mathbf{z}$ and the augmented representation $\mathbf{z}_{\text{aug}}$.
An augmentation is considered successful if:
\begin{eqnarray}
	\text{true} &=& \mathbf{1}\left[\cos(\mathbf{z}, \mathbf{z}_{\text{aug}}) > \alpha\right],
\end{eqnarray}
where $\alpha$ is a learnable threshold. This design encourages the generation of informative multi-view pairs for contrastive learning.

In the fine-tuning phase, where labeled data is used to optimize the model for downstream tasks, we measure augmentation success by the classification accuracy on the augmented samples. Specifically, an augmentation is successful if:
\begin{eqnarray}
	\text{true} &=& \mathbf{1}\left[\frac{1}{|B_{\text{aug}}|} \sum_{i \in B_{\text{aug}}} \mathbf{1}[\hat{y}_i = y_i] > \beta\right],
\end{eqnarray}
where $\beta$ is another learnable threshold, and $B_{\text{aug}}$ denotes the batch of augmented samples. This encourages the model to prioritize augmentations that help improve generalization and robustness during fine-tuning.

\subsection{Contrastive Learning Pre-training}
EEG signals encompass abundant temporal information as well as critical frequency-domain characteristics.
Due to the diversity introduced by data augmentation, considerable variations of the same signal may emerge from different views.
Consequently, maintaining consistency of features across these augmented views while effectively distinguishing between distinct samples constitutes a pivotal task during the pre-training stage.
\subsubsection{Design of Time-frequency Encoder}
To fully exploit the multi-dimensional characteristics of EEG signals, we design a CNN-based encoder consisting of three complementary branches that extract temporal, frequency, and time-frequency representations.
Given an input signal \(X \in \mathbb{R}^{C \times T}\), where \(C\) is the number of EEG channels and \(T\) is the signal length, the temporal encoder applies stacked 1D convolutions to extract local and sequential patterns.
Each layer performs the following operation:
\begin{eqnarray}
	H^{(l)} &=& \mathrm{ReLU}\bigl(\mathrm{BN}(\mathrm{Conv1D}(H^{(l-1)}))\bigr).
\end{eqnarray}

To capture spectral features, the frequency encoder applies Fast Fourier Transform (FFT) to the input signal \(X\), resulting in its complex-valued frequency representation \(\mathcal{F}(X)\).
The encoder then computes the magnitude spectrum \(|\mathcal{F}(X)|\), which reflects the amplitude of each frequency component while discarding phase information.
This representation serves as the basis for extracting frequency-domain features from the input.

To jointly model temporal and spectral information, the time-frequency encoder concatenates the raw signal with its magnitude spectrum along the channel dimension and processes the resulting hybrid representation via a convolutional encoder:
\begin{eqnarray}
	H_{tf} &=& \mathrm{Encoder}\bigl([X; |\mathcal{F}(X)|]\bigr).
\end{eqnarray}
After obtaining the temporal feature \(h_t\), frequency feature \(h_f\), and time-frequency joint feature \(h_{tf}\), we apply an independent projection head to each, implemented as a two-layer fully connected network:
\begin{eqnarray}
	z &=& W_2\left(\mathrm{ReLU}(W_1 h + b_1)\right) + b_2,
\end{eqnarray}
where \(h\) is the input feature from each branch, and \(W_1\), \(W_2\), \(b_1\), and \(b_2\) are learnable parameters.

This design ensures that the distinctiveness of each representation is preserved during projection, providing a more discriminative feature space for multi-view contrastive learning and improving the generalization and robustness of downstream tasks.

\subsubsection{Design of Contrastive Loss Function}
To ensure that the feature representations of the same signal remain consistent across different augmented views while effectively distinguishing different samples, we adopts a normalized temperature-scaled cross entropy loss function.
This loss function measures feature similarity across different views using normalized cosine similarity and employs a temperature parameter \(\tau\) to scale the similarity, balancing the contrast between positive and negative samples.

Given a batch of \(N\) samples, the normalized projection vectors of the same sample \(k\) under two different views \(i\) and \(j\) are denoted as \(\hat{\mathbf{z}}_k^{(i)}\) and \(\hat{\mathbf{z}}_k^{(j)}\), respectively.
Each projection vector is normalized to have a unit norm, which ensures numerical stability when computing cosine similarity.
The cosine similarity between the two projections is computed as the dot product of the normalized vectors, scaled by a temperature hyperparameter \(\tau\).
The temperature \(\tau\) controls the concentration level of the similarity distribution, affecting how sharply the model distinguishes between similar and dissimilar pairs.
For sample \(k\) in view \((i,j)\), the corresponding NT-Xent loss is defined as:
\begin{eqnarray}
	\ell_{k}^{(i,j)} &=& - \log \frac{\exp\bigl(\text{sim}(\hat{\mathbf{z}}_k^{(i)}, \hat{\mathbf{z}}_k^{(j)})\bigr)}{\sum_{i=1}^{N} \exp\bigl(\text{sim}(\hat{\mathbf{z}}_k^{(i)}, \hat{\mathbf{z}}_l^{(j)})\bigr)}.
\end{eqnarray}
The denominator aggregates the similarities between sample \(k\) in view \(i\) and all samples in view \(j\) within the batch, thereby treating different views of the same signal as positive pairs while considering different signals as negative pairs.

To ensure symmetry, an additional loss term \(\ell_{k}^{(j,i)}\) is defined for sample \(k\) in view \(j\). The final contrastive loss is obtained by averaging over all view pairs:
\begin{eqnarray}
	\mathcal{L}_{\text{contrast}}=\frac{2}{M(M-1)} \sum_{i=1}^{M-1} \sum_{j=i+1}^{M} \frac{1}{N} \sum_{k=1}^{N} \left( \ell_{k}^{(i,j)} + \ell_{k}^{(j,i)} \right),
\end{eqnarray}
where \(M\) denotes the total number of views.

This loss function design ensures that the same signal maintains consistent representations under different augmentations while significantly enhancing the model’s ability to differentiate between distinct signals, thereby providing a robust and discriminative feature foundation for downstream tasks.

\subsection{Fine-tuning for Cross-dataset Scenario}
To enable the PD detection model to adapt to different datasets, we perform fine-tuning using a small amount of labeled data.
Similar to the augmentation strategy employed in the pre-training phase, we initialize a dynamic augmentation manager prior to fine-tuning.
This module augments the limited labeled data and dynamically adjusts the augmentation strategy in real-time.
Such a design not only reduces the reliance on labeled samples but also enhances the generalization capability of the finetuned model.
Subsequently, we incorporate multiple strategies to further optimize the model during fine-tuning. 

\subsubsection{Layer-wise Unfreezing and Progressive Training}
We adopt a top-down layer-wise unfreezing strategy, where layers are progressively unfrozen with decayed learning rates:
\begin{eqnarray}
	\eta_i^{(l)} &=& \eta_0 \cdot \gamma^{L - l}.
\end{eqnarray}
At the stage \(i\), the set of unfrozen layers is defined as:
\(\Omega_i = \{L, L-1, \ldots, L-(i-1)\}\).
This allows higher layers to adapt quickly to downstream tasks while preserving low-level pretrained features and preventing catastrophic forgetting.

\subsubsection{Regularization Techniques}

To further enhance generalization during fine-tuning and mitigate overfitting, we incorporate the following regularization techniques:

\textbf{Label smoothing:}
Using strict one-hot labels can cause the model to become overly confident in its predictions, which may lead to overfitting.
To address this, we apply label smoothing by slightly softening the target labels.
Instead of assigning full probability to the correct class and zero to all others, a small portion of the probability mass is distributed evenly among the incorrect classes.
This technique reduces overconfidence, encourages the model to produce more calibrated and distributed probability outputs, and ultimately enhances generalization performance.

\textbf{Weight decay:}
To control model complexity and prevent overfitting, L2 regularization is applied in the loss function.
This helps constrain parameter magnitudes, leading to improved generalization.

\subsubsection{Optimization Strategy}

We adopt Lookahead optimizer, which maintains two sets of weights: fast weights \(\theta\) and slow weights \(\phi\), enabling stable updates. The update rules are:
\begin{eqnarray}
	\theta_{t+1} &=& \theta_t - \alpha \nabla f(\theta_t) \\
	\phi_{t+k} &=& \phi_t + \beta \bigl(\theta_{t+k} - \phi_t\bigr),
\end{eqnarray}
where \(\alpha\) is the learning rate of the inner optimizer, \(\beta\) is the slow weight update rate, and \(k\) is the number of inner updates.
This mechanism smooths parameter updating and improves training stability.

In the later training stages, we apply SWA by averaging the model weights from several checkpoints.
This technique smooths the optimization path and reduces the model’s sensitivity to sharp local minima, leading to better generalization.

\section{Experiments}

\subsection{Datasets}
We use three publicly available datasets UI dataset \cite{c:22}, UNM dataset \cite{c:23} and UC dataset \cite{c:24}.
UI includes 28 participants (14 PD patients and 14 healthy controls), and UNM includes 54 participants (27 PD patients and 27 healthy controls).
All participants are off levodopa medication for at least 12 hours before data collection, and the data are collected in a resting state with eyes open.
EEG data are recorded using a 64-channel Brain Vision system, with a frequency range of 0.1-100 Hz and a sampling rate of 500 Hz, acquired through PRED+CT platform.
UC dataset includes 15 right-handed PD patients (8 females, mean age 62.6 ± 8.3 years), and 16 matched HC (9 females, 63.5 ± 9.6 years) based on age, gender, and handedness.
The patients are recruited from Scripps Clinic in La Jolla, California, and HC are volunteers from the local community.

In data preprocessing, EEG time series from all non-reference electrodes are used.
To unify referencing, the UI and UC datasets are re-referenced to the CPz channel following the international 10–20 system.
Electrodes not shared between datasets are removed, yielding 30 consistent non-reference channels. Signals are band-pass filtered between 1–45\,Hz using a Hamming window filter. Each session is segmented into non-overlapping 5-second epochs (2,500 points per channel) and normalized via z-score.
Each sample is formatted as a matrix of 8 batches, 60 channels, and 2,500 time points, with binary labels (1 for PD, 0 for controls).
These steps ensure consistent and reliable data for downstream analysis.


\subsection{Settings}
During pre-training, EEG data are segmented and augmented with Gaussian noise (mean 0, std 0.1), temporal shift (max 50), amplitude scaling (0.8–1.2), random masking (length 10), frequency shift (max 2), spectral scaling (0.5–1.5), and band-specific noise (sampling rate 500\,Hz).
A dynamic sampling strategy is used with initial temperature 1.0 and success threshold $\alpha = 0.5$.
A CNN-based time-frequency encoder is trained for 30 epochs (batch size 32) using AdamW (learning rate $1 \times 10^{-4}$) and CosineAnnealingWarmRestarts (cycle length 10, multiplier 2).
NT-Xent loss is applied with temperature 0.1, and early stopping is triggered if validation loss does not improve by $1 \times 10^{-4}$ within 10 epochs.
All optimal parameters are selected from multiple experiments.
During fine-tuning, the success threshold \(\beta\) is also set to 0.5.
We use 90\% of the data as the fine-tuning test set and 10\% of the data as the training set based on the subject IDs, and further divide the training set into 50\% or 90\% as the validation set.
The batch size remains 32 and the number of training epochs is set to 300.
The learning rate is increased to \(1 \times 10^{-3}\), and the weight decay is set to \(1 \times 10^{-4}\).
To reduce overfitting caused by limited data, label smoothing is applied to the cross-entropy loss with a smoothing factor of 0.1.
Similarly, all optimal parameters are selected from multiple experiments.

\subsection{Comparative Results}

\subsubsection{Level of different cross-datasets}
\begin{table}[ht]
	\centering
	\renewcommand{\arraystretch}{1.7} 
	\caption{Comparison of the effectiveness with existing methods.}
	\label{tab:comparison}
	\begin{tabular}{p{30pt}p{80pt}p{15pt}p{15pt}p{15pt}p{15pt}}
		\specialrule{1pt}{0pt}{0pt}
		\hline
		\textbf{Work} & \textbf{Method} & \textbf{Train} & \textbf{Test} & \textbf{ACC} &\textbf{F1}\\
		\specialrule{1pt}{0pt}{0pt}
		\hline
		Sugden et al.\cite{c:8} & Channel-wise convolutional neural network & UNM & UI & 0.828 & 0.841\\
		Anjum et al.\cite{c:28} & Linear-predictive-coding EEG Algorithm & UNM & UI & 0.857 & 0.856\\
		\addlinespace[3pt] 
		Avvaru et al.\cite{c:29} & Frequency-domain convergent cross-mapping & UNM & UC & 0.839 & 0.840$^*$\\
		\addlinespace[3pt] 
		\multirow{2}{=}{Qiu et al.\cite{c:5}} & \multirow{2}{=}{Multiscale convolutional prototype network} & UC & UNM & 0.902 & 0.901$^*$\\
		& & UNM & UC & 0.865 & 0.864$^*$\\
		\hline
		\multirow{6}{=}{ours} & \multirow{6}{=}{Unsupervised +5\%fine-tuning} & UNM & UI & \textbf{0.971} & \textbf{0.971}\\
		& & UNM & UC & \textbf{0.873} & \textbf{0.873}\\
		& & UI & UNM & \underline{0.893} & \underline{0.891}\\
		& & UI & UC & 0.846 & 0.847\\
		& & UC & UNM & 0.873 & 0.872\\
		& & UC & UI & 0.963 & 0.963\\

		\hline
		\specialrule{1pt}{0pt}{0pt}
	\end{tabular}
	\vspace{2mm}
	\begin{tablenotes}
		\item $^*$ Estimated from reported accuracy assuming class balance.
	\end{tablenotes}
\end{table}

In the cross-dataset comparison experiments, we focus on evaluating the model's performance across multiple target datasets to assess its generalization capability to unseen data.
The experimental results demonstrate that existing methods exhibit inconsistent performance across different testing scenarios.
Specifically, the channel-wise convolutional neural network proposed by Sugden et al. \cite{c:6}, trained on the UNM dataset and directly tested on the UI dataset, achieves an accuracy of 0.828 and an F1-score of 0.841.
Similarly, Anjum et al. \cite{c:28} employ a linear predictive coding (LPC) approach, attaining an accuracy of 0.857 and an F1-score of 0.856 under the same train-test configuration.
However, their method heavily relies on handcrafted feature extraction.
Avvaru et al. \cite{c:29} introduce a frequency-domain convergent cross mapping method, which yields an accuracy of 0.839 when trained on the UNM dataset and tested on the UC dataset.
These three methods do not report results on other cross-dataset settings, suggesting potential limitations in their generalization ability.
In contrast, the multiscale convolutional prototype network proposed by Qiu et al. \cite{c:5} is evaluated across two datasets, achieving accuracies of 0.902 and 0.865 on the UNM and UC test sets, respectively. Nevertheless, their method requires fully labeled data in the source domain and partially labeled data in the target domain for training.

In comparison, our proposed framework leverages entirely unlabeled data during the pre-training phase and requires only 5\% of labeled data for fine-tuning.
The results show that our method achieves the state-of-the-art (SOTA) performance in all the three cross-dataset scenarios, with accuracies and F1-scores of 0.971 on the UI dataset, 0.873 on the UC dataset, and 0.893 on the UNM dataset.
These results demonstrate that our framework effectively captures domain-invariant EEG representations, thereby significantly enhancing generalization performance and robustness in cross-dataset EEG-based classification tasks.

\subsubsection{Effect of Our Pretrained Model}

\begin{table}[h]
	\centering
	\renewcommand{\arraystretch}{1.3} 
	\setlength{\tabcolsep}{11pt} 
	\caption{Performance comparison of different self-supervised learning models on the UI and UC test sets.}
	\label{tab:ssl_comparison}
	\begin{tabular}{lcccc}
		\specialrule{1pt}{0pt}{0pt}
		\toprule
		\multirow{2}{*}{\textbf{Models}} & \multicolumn{2}{c}{\textbf{UI Test}} & \multicolumn{2}{c}{\textbf{UC Test}} \\
		\cmidrule(lr){2-3} \cmidrule(lr){4-5}
		& \textbf{ACC} & \textbf{F1} & \textbf{ACC} & \textbf{F1} \\
		\specialrule{1pt}{0pt}{0pt}
		\midrule
		SimCLR \cite{c:10} & 0.559 & 0.451 & 0.539 & 0.431 \\
		MoCo \cite{c:25} & 0.636 & 0.621 & 0.617 & 0.609 \\
		BYOL \cite{c:26} & 0.678 & 0.676 & 0.656 & 0.643 \\
		Barlow Twins \cite{c:27} & 0.701 & 0.716 & 0.695 & 0.708 \\
		\textbf{Ours} & \textbf{0.773} & \textbf{0.773} & \textbf{0.757} & \textbf{0.766} \\
		\specialrule{1pt}{0pt}{0pt}
		\bottomrule
	\end{tabular}
\end{table}

To evaluate the effectiveness of our proposed contrastive learning pre-training method, we compare it with four state-of-the-art pre-training approaches on the UI and UC datasets.

Since the goal is to assess the efficacy of pre-training in EEG representation learning, we conduct cross-dataset evaluation without fine-tuning, and the results are presented in Table~\ref{tab:ssl_comparison}.
SimCLR \cite{c:10} relies on extensive data augmentation and cross-entropy loss to bring positive pairs closer in the representation space.
However, given the limited number of EEG samples and high noise levels, it struggles to capture fine-grained differences in contrastive signals.
MoCo \cite{c:25} mitigates the small batch size constraint by maintaining a momentum encoder and a queue of negative samples, enabling more diverse negative samples for contrastive learning.
Nevertheless, the inherent randomness in negative sample selection limits the improvement in representation quality.
BYOL \cite{c:26} eliminates the reliance on negative samples and instead leverages a self-supervised learning paradigm, yielding more stable representations.
However, direct self-supervision remains constrained when handling the complex temporal and spectral characteristics of EEG signals.
Barlow Twins \cite{c:27} introduces a redundancy reduction strategy by enforcing the cross-correlation matrix of learned representations to approximate an identity matrix, effectively reducing information redundancy and enhancing representation discriminability.

Building upon these prior methods, we incorporate a dynamic data augmentation strategy and a specially designed time-frequency joint encoder.
This approach ensures consistency in feature representations across different augmented views while fully leveraging both temporal and spectral information in EEG signals.
Experimental results demonstrate that our method consistently achieves the best performance on both test sets (UI Test ACC: 0.7727, UC Test ACC: 0.7569).
These findings indicate that our approach effectively enhances model generalization and robustness, making it well-suited for fine-tuning on downstream datasets.

\subsection{Ablation Study}

\begin{table}[ht]
	\centering
	\renewcommand{\arraystretch}{1.65} 
	\setlength{\tabcolsep}{15pt} 
	\caption{Comparison of pre-training and fine-tuning ablation modules}
	\label{tab:ablation}
	\begin{tabular}{ccc} 
		\hline
		\specialrule{1pt}{0pt}{0pt}
		\multicolumn{3}{l}{\textit{\textbf{pre-training}}} \\ 
		\hline
		\textbf{DA Module} & \textbf{TFF Module} & \textbf{Performance (ACC)} \\
		$\times$ & $\checkmark$ & UI: 0.697, UC: 0.653 \\
		$\checkmark$ & $\times$ & UI: 0.746, UC: 0.721 \\
		$\checkmark$& $\checkmark$ & \textbf{UI: 0.772}, \textbf{UC: 0.756} \\
		\hline
		\specialrule{1pt}{0pt}{0pt}
		\multicolumn{3}{l}{\textit{\textbf{Fine-tuning}}} \\ 
		\hline
		\multicolumn{2}{l}{Unsupervised} & UI: 0.772, UC: 0.756 \\
		\multicolumn{2}{l}{Unsupervised + 1\% labeled data} & UI: 0.915, UC: 0.816 \\
		\multicolumn{2}{l}{Unsupervised + 5\% labeled data} & \textbf{UI: 0.971}, \textbf{UC: 0.873} \\
		\hline
		\specialrule{1pt}{0pt}{0pt}
	\end{tabular}
\end{table}

To validate the effectiveness of each proposed model, we conduct an ablation study.
First, we select the UNM dataset for contrastive learning pre-training.
We then evaluate the trained models on the UI and UC datasets under cross-dataset testing conditions while removing either the dynamic data augmentation strategy (DA Module) or the time-frequency information fusion (TFF Module).
The results are presented in Table~\ref{tab:ablation}.
When only the time-frequency fusion module is retained, the accuracy on the UI and UC datasets is 0.697 and 0.653, respectively.
Conversely, when only the dynamic data augmentation module is preserved, the accuracy improved to 0.746 and 0.721.
These results demonstrate that the combination of both modules in our pre-training framework significantly enhances cross-dataset generalization, achieving accuracy of 0.772 and 0.756 on the UI and UC datasets, respectively.
This highlights the superiority of our proposed method.

During the fine-tuning phase, we utilize 1\% and 5\% of labeled data to fine-tune the pretrained model. With only 1\% of labeled data, the model achieves cross-dataset accuracy of 0.915 and 0.816 on the UI and UC datasets, respectively. When increasing the labeled data to 5\%, the accuracy further improved to 0.971 and 0.873.
These results demonstrate that fine-tuning with a small amount of labeled data significantly enhances the model’s classification performance on the corresponding downstream datasets.
\begin{figure*}[t]
	\centering
	\includegraphics[width=1\textwidth]{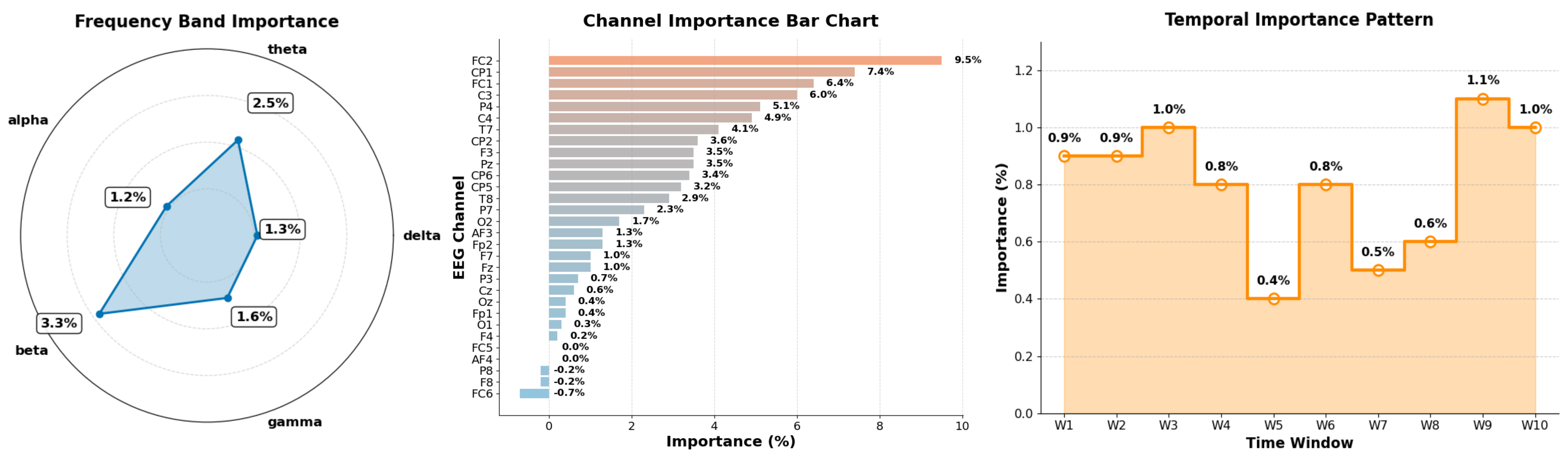}
	\caption{The results of the interpretability experiment are shown as the importance of frequency band (left), channel (middle), and time segment (right).}
	\label{fig2}
\end{figure*}
\begin{figure}[t]
	\centering
	\includegraphics[width=0.49\textwidth]{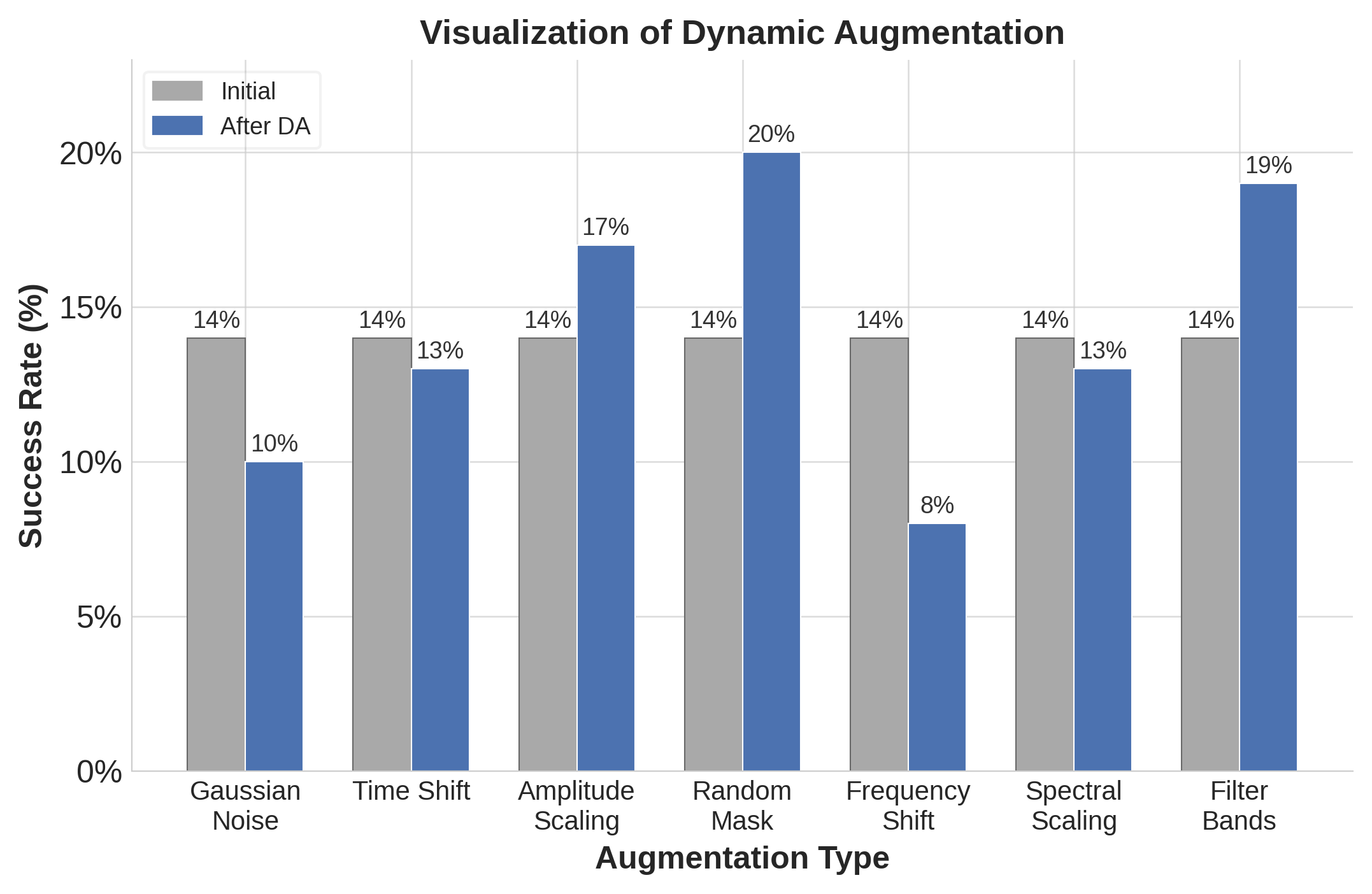}
	\caption{Visualization of the weight change of the data augmentation.}
	\label{fig3}
\end{figure}
\subsection{Interpretability Experiment}

To elucidate the decision-making mechanisms of deep learning models in identifying PD from EEG signals, we conduct a perturbation-based multi-dimensional interpretability analysis.
This approach quantifies the importance of different features to model predictions by selectively removing specific features and observing the corresponding performance degradation. We analyze EEG signal characteristics from three perspectives.

\textbf{Frequency band importance: }The signal is transformed into the frequency domain using the Fast Fourier Transform (FFT), and individual frequency bands (delta: 0.5–4 Hz, theta: 4–8 Hz, alpha: 8–13 Hz, beta: 13–30 Hz, gamma: 30–100 Hz) are sequentially removed to assess the impact on classification accuracy.  

\textbf{Channel importance:} EEG channels are individually masked by setting their values to zero, allowing evaluation of the contribution of different brain regions to prediction.  

\textbf{Temporal window importance:} The signal is divided into 10 equal time windows, with each window sequentially occluded to analyze the distribution of critical temporal features.

Figure \ref{fig2} presents the results of the interpretability experiment.
From left to right, the three subfigures illustrate the importance of different frequency bands, channels, and temporal segments in disease detection.
The leftmost subfigure indicates that the beta frequency band is the most critical for disease recognition.
This observation is consistent with existing studies, which demonstrates that dopamine depletion in PD leads to an abnormal increase in $\beta$-band neural oscillations in the basal ganglia circuit, supporting the validity of our experiment \cite{c:30,c:31,c:32,c:33}.
The middle subfigure highlights the contribution of specific EEG channels, with FC2 (8.5\%), CP1 (7.4\%), and FC1 (6.4\%) exhibiting the highest signal importance.
The rightmost subfigure shows that features in both early and late temporal segments (W9: 1.1\%, W1: 0.9\%) are more prominent.
The significance of specific channels and temporal windows in disease detection remains largely unexplored in neuroscience. Therefore, our findings offer potential new biomarkers for the development of non-invasive early diagnosis and monitoring tools in clinical practice, facilitating the optimization of therapeutic intervention timing.

\subsection{Visualization of Weight Changes of DA}

The proposed dynamic data augmentation module adaptively adjusts the weights and composition strategies of various augmentation techniques throughout the training process.
To investigate its behavior, we visualize the effectiveness of seven major augmentation methods by tracking their success rate variations under dynamic adjustments in a representative experiment.
As shown in Figure \ref{fig3}, the success rates of random masking and band-pass filtering increased from the initial 14\% to 20\% and 19\%, respectively.
Amplitude scaling improved to 17\%, while time shifting and spectral scaling slightly decreased to 13\%.
In contrast, Gaussian noise injection and frequency shifting dropped to 10\% and 8\%, respectively.

These results are intended to illustrate the variation in adaptation across different augmentation strategies within the dynamic framework, rather than implying that the model relies solely on a few high-weighted methods.
In practice, the module continuously adjusts augmentation strategies in response to training progress and data characteristics, thereby providing an optimized data distribution to facilitate robust model learning.

\section{Conclusion and Future Work}
This paper presents MCLPD, an EEG-based Parkinson’s disease detection method that integrates multi-view contrastive learning with advanced fine-tuning strategies.
By generating diverse time-frequency augmented views, MCLPD learns robust and discriminative representations from limited labeled data.
Fine-tuning incorporates label smoothing, Lookahead optimizer, SWA, and other techniques to enhance generalization and stability.
Experiments show that MCLPD significantly outperforms supervised baselines in accuracy and F1-score, demonstrating the effectiveness of the proposed framework.

Future work will focus on improving computational efficiency for real-time deployment on low-power devices.
Moreover, to address individual variability in EEG signals, future directions include personalized modeling, cross-subject transfer learning, and federated learning to enhance generalization across subjects.

\clearpage
\begin{ack}
The work of Qian Zhang is partially supported by Intelligent Policing Key Laboratory of Sichuan Province No. ZNJW2024KFQN006.
The work of Zhe Wang is partially supported by Natural Science Foundation of China under Grant No. 62476087,
Shanghai Municipal Education Commission's Initiative on Artificial Intelligence-Driven Reform of Scientific Research Paradigms and Empowerment of Discipline Leapfrogging, 
National Key Research and Development Program of China under Grant 2022YFB3203500.

\end{ack}
\bibliography{mybibfile.bib}

\end{document}